\documentclass[11pt]{article}

\usepackage[final]{acl}

\usepackage{times}
\usepackage{latexsym}
\usepackage[T1]{fontenc}
\usepackage[utf8]{inputenc}
\usepackage{microtype}
\usepackage{inconsolata}
\usepackage{graphicx}
\usepackage{booktabs}
\usepackage{multirow}
\usepackage{caption}
\usepackage[english]{babel}

\newcommand{\apm}[2]{\textnormal{#1\textsubscript{\textpm#2}}}

\title{From Abusive Language Classification to Sequence Labeling Identification}

\author{Nicolas Zampieri$^*$, \textbf{Ignacio Lopez$^\dagger$}, \textbf{Manon Girard$^*$}, \textbf{Jeremy Auguste$^*$} \\
  Everdian / Marseille, France \\
  $^*$\texttt{firstname.lastname@everdian.com}, $^\dagger$\texttt{ignlopezca@gmail.com}}
\begin{document}
\maketitle

\begin{abstract}
Industrial content moderation must process massive message streams under tight latency constraints, yet most abusive language (AL) detection systems rely on sentence-level classification (ALC), which neither localizes abusive spans nor identifies who is targeted.
We define \textbf{Abusive Language Identification} (ALI) as a sequence-labeling task that jointly extracts AL spans and target mentions, and assess whether this approach can be used for text moderation.
On a pilot corpus drawn from a production moderation pipeline, we compare ALI with ALC on cross-domain generalization and implicit abuse, and we also evaluate AL and target span detection.
ALI remains competitive with ALC while providing localized outputs for moderators, with a modest and configuration-sensitive advantage on implicit abuse.
Exact AL boundaries and target spans remain difficult to recover.
We complement this comparison with a qualitative analysis and discuss perspectives on complete target--span linking and on structured benchmarks for ALI.
\end{abstract}

% !TEX root = acl_latex_en.tex
\section{Introduction}

\begin{small}
    \textbf{Warning:} \emph{This paper contains offensive content and may be distressing.}
\end{small}

Industrial content moderation operates under severe throughput constraints: major social media platforms and thematic forums generate volumes of text that make exhaustive human review impossible\footnote{2025 estimates put the number of short daily messages on major micro-blogging platforms at over 500 million, and daily contributions on community forums at several million.}.
Operators must process massive streams of messages within tight latency budgets, while still identifying content that may require differentiated action, from standard removal to priority escalation and protective follow-up~\cite{guillen2023}.
In this context, detection systems must deliver not only speed and scale, but outputs that are directly actionable for analysts who are not experts.

Most research on abusive language (AL) detection remains framed as Abusive Language Classification (ALC)~\cite{basile-etal-2019-semeval, zampieri-etal-2019-semeval, rottger-etal-2021-hatecheck}.
ALC assigns a single label to an entire message, but does not localize which spans constitute AL nor explicitly identify targeted entities.
This limits operational usefulness when a single message contains multiple abusive instances, when only part of a text is harmful, or when the moderation response depends on \textit{who} is targeted~\cite{10.1145/3366424.3385778, pandya-etal-2025-hostility}.

Recent work has moved toward finer-grained annotation schemes.
\citet{zampieri-etal-2023-target} introduced target-based offensive (TBO) language, bridging sentence-level harmfulness labels with token-level annotations of offensive arguments and their targets.
JL-Hate~\cite{buyukdemirci-etal-2024-jl} further proposes joint token-level detection of hate/offensive spans and target entities.
Inspired by these works, we propose the \textbf{Abusive Language Identification} (ALI) task as a native sequence labeling formulation that jointly extracts AL spans and target mentions, together with explicit links between them.

In this paper, we design a \textbf{pilot corpus} for ALI, which covers all categories of AL, from abusive terms and offensive speech to threats and hate speech, as well as other forms of abusive language encountered in industrial moderation.
We evaluate ALI in out-of-domain settings due to the lack of a dedicated ALI dataset. 
The main purpose of this paper is to assess whether an ALI-based approach can be used for text moderation.
To this end, we compare ALI and ALC along two axes.
We first focus on \textbf{cross-domain generalization}, as models trained on one annotation scheme or platform rarely remain robust on unseen corpora~\cite{karan-snajder-2018-cross, 10.1145/3331184.3331262, yin2021towards}.
We also examine \textbf{implicit AL}\footnote{The detailed implicit and explicit AL analysis is reported in Appendix~\ref{annex:abusive_language_implicit_explicit} because of paper-length limits.}, where harmful intent is conveyed without overt slurs and remains particularly difficult to detect~\cite{wiegand-etal-2021-implicitly-abusive, sap-etal-2020-social, elsherief-etal-2021-latent}.
We hypothesize that ALI better captures such cases than ALC by localizing relational cues in context, such as pairing a seemingly benign phrase with a specific addressee, rather than relying on isolated lexical signals at the message level.
We also assess AL and target span detection, since a moderation decision depends on localizing the abusive content and identifying who is targeted.

Our results indicate that ALI remains competitive with ALC in cross-domain settings while offering natively interpretable span-level outputs for moderators, and may offer some advantage on implicit abuse, where contextual cues matter more than isolated lexical signals, although gains are modest and configuration-sensitive.
Span detection itself remains challenging: exact AL boundaries are difficult to recover and target spans are likewise hard to delimit.
We conclude that sequence labeling is a viable alternative to ALC for industrial moderation, and discuss perspectives on complete target-span linking and on structured benchmarks for ALI.

\section{Related work}

\textbf{AL detection.}
Most research on AL detection explores the problem as an ALC task~\cite{waseem-hovy-2016-hateful, davidson2017icwsm, Founta_Djouvas_Chatzakou_Leontiadis_Blackburn_Stringhini_Vakali_Sirivianos_Kourtellis_2018, basile-etal-2019-semeval, zampieri-etal-2019-semeval}.
Shared tasks and evaluation benchmarks have consolidated this paradigm~\cite{basile-etal-2019-semeval, rottger-etal-2021-hatecheck}, while surveys highlight persistent challenges in definition, bias, and evaluation~\cite{yin2021towards, fortuna-etal-2022-directions, JAHAN2023126232}.
Transformer-based encoders pretrained or fine-tuned on AL corpora remain the dominant modeling choice~\cite{caselli-etal-2020-feel, caselli2021hatebertretrainingbertabusive, sarkar2021fbertneuraltransformeridentifying, awal2021angrybertjointlearningtarget}.
Although effective for screening, ALC provides limited actionable structure: it neither localizes AL spans nor explicitly highlights targeted entities, limiting its usefulness for operational moderation workflows that require differentiated responses~\cite{guillen2023}.

\textbf{Span-level sequence labeling.}
Beyond ALC, an increasing number of corpora adopt token-level sequence labeling to localize harmful content directly in text.
The SemEval-2021 shared task on toxic span detection~\cite{pavlopoulos-etal-2021-semeval} formalized this paradigm by training models to identify the minimal textual units responsible for a toxicity label.
HateXplain ~\cite{Mathew_Saha_Yimam_Biemann_Goyal_Mukherjee_2021} also provides token-level labels, but these annotations remain secondary to the sentence-level class and are not designed as a standalone identification task.
More recently, LLM-based pipelines have been proposed to produce span-level labels~\cite{nirmal-etal-2024-towards}, yet still derive spans from a predefined classification decision rather than from native structured annotation.
Across these resources, labeled spans are typically very short (often single tokens), omit target identification, and do not cover a unified taxonomy of abuse types (threats, hate speech, non-targeted profanity).

\textbf{Target-aware AL resources.}
Identifying \textit{who} is attacked is increasingly recognized as essential, especially for threats, hostility, and hate speech directed at individuals or groups~\cite{10.1145/3366424.3385778, vidgen-etal-2021-introducing, pandya-etal-2025-hostility}.
Recent corpora jointly annotate aggressiveness and target categories at the span level~\cite{damo-etal-2025-hodiat}, reflecting growing interest in structured, target-aware sequence labeling beyond sentence-level harmfulness.
\citet{barbarestani-etal-2022-annotating} extend HX with span-level target communities, but these target annotations are not explicitly linked to the corresponding AL spans.
JL-Hate~\cite{buyukdemirci-etal-2024-jl} proposes joint learning of hate/offensive spans and target entities in English and Turkish, yet does not model explicit associations between an AL span and its target mention.
\citet{zampieri-etal-2023-target} introduce target-based offensive (TBO) language, annotating offensive arguments together with their targets and bridging sentence-level harmfulness with token-level structure.
\citet{atil2026trust} introduce {TR}u{ST}, which merges prior social-media and LLM-generated resources under a unified scheme for binary toxicity, a social-group target (8 coarse / 24 fine labels), and toxic spans, with a human-annotated subset of about 11k examples.
% On that subset, fine-tuned PLMs outperform zero-shot LLMs and moderation APIs (toxicity F1 $0.79$ for {R}o{BERT}a; span F1 $0.70$ for {S}pan{BERT}; fine-grained target F1 $0.72$ for {BERT}).
Unlike JL-Hate and TBO, {TR}u{ST} targets are categorical group labels rather than token-level mentions.

\textbf{Cross-domain generalization.}
A recurring finding is that ALC models generalize poorly across corpora and platforms~\cite{karan-snajder-2018-cross, 10.1145/3331184.3331262, caselli-etal-2020-feel, yin2021towards, antypas-camacho-collados-2023-robust}.
Domain adaptation via unsupervised fine-tuning on the target distribution~\cite{bose-etal-2021-unsupervised} and multi-task training across heterogeneous datasets~\cite{YUAN2025101690} partially mitigate this issue, but performance remains sensitive to annotation scope and label definition mismatches.
These observations motivate our cross-domain evaluation comparing ALI and ALC.

\textbf{Implicit AL.}
Implicit abuse, where harmful intent is conveyed without overt slurs, remains a major bottleneck~\cite{sap-etal-2020-social, wiegand-etal-2021-implicitly-abusive, elsherief-etal-2021-latent}.
The Implicit Hate Corpus (IHC)~\cite{elsherief-etal-2021-latent} explicitly separates implicit and explicit hateful messages, showing ties between implicit abuse and stereotypical knowledge.
Models trained without such phenomena struggle to detect them at inference time~\cite{yin2021towards, caselli-etal-2020-feel}.
Recent studies continue to stress the difficulty of implicit formulations in web and conversational settings~\cite{boudraa-etal-2025-implicit, carvallo-etal-2025-hate}.
Span-level modeling is a plausible response to this challenge: localizing abuse and its target may capture relational cues (e.g., pairing a benign-looking phrase with a specific addressee) that ALC models miss~\cite{wiegand-etal-2021-implicitly-abusive}.
Building on prior works, we deliberately include implicit discourse in our pilot corpus and provide an evaluation on this type of AL.

\textbf{LLMs for AL detection.}
LLMs have been explored for AL detection through fine-tuning~\cite{sen2024hatetinyllmhatespeech}, prompting~\cite{ghorbanpour-etal-2025-prompting}, retrieval augmentation~\cite{mnassri-etal-2025-rag}, and synthetic data generation~\cite{schmidhuber2024llm}.
While promising on benchmark tasks, their computational and environmental cost remains a barrier for large-scale, low-latency moderation pipelines~\cite{fortuna-etal-2022-directions}.
Our experiments therefore focus on encoder-based sequence labeling models, which offer more deployable trade-off between accuracy and inference cost while delivering structured, interpretable outputs.

% !TEX root = acl_latex_en.tex
\section{Proposed methodology}
\label{sec:methodology}
In this section, we detail the methodology of our ALI approach, designed to meet the constraints of large-scale analysis. 
Our goal is to produce structured predictions, AL spans associated with their targets, offering better readability for operators who are not NLP experts. 
Unlike ALC approaches that provide a simple binary label, our methodology draws on work in targeted sentiment analysis~\cite{orbach_yaso_2021} and aims to locate problematic elements precisely in the text as proposed by \citet{zampieri-etal-2023-target}.
 
\begin{figure}
    \centering
    \includegraphics[width=.95\columnwidth]{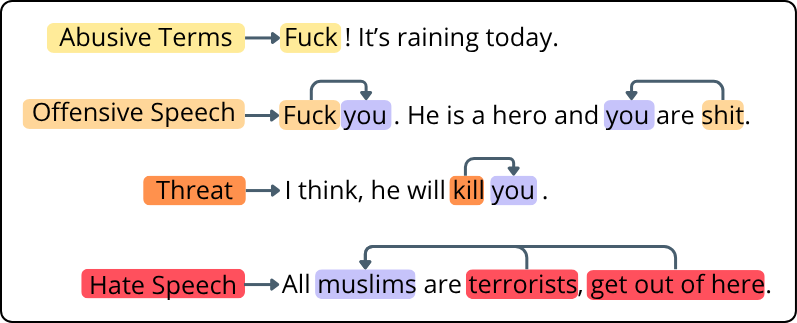}
    \captionof{figure}{Annotation examples for each class: \textit{Abusive terms}, \textit{Offensive speech}, \textit{Threat}, and \textit{Hate speech}. 
    Text highlighted in blue designates the target, and other highlighted text designates AL (shown here to the left of the sentences). The arrow represents the association between AL and its target.}
    \label{tagging_samples}
\end{figure}

We present here the full definition of the ALI task, including the taxonomy and annotation protocol; although our full methodology integrates explicit linking between AL spans and their targets (arrows in Figure~\ref{tagging_samples}), the empirical evaluation focuses on the first step, detecting AL spans, to establish a rigorous comparison with ALC approaches.
% OLD: ...while noting that the empirical evaluation in this article focuses primarily on detecting AL spans (first step), leaving fine-grained evaluation of target--span linking to future work.
Complete evaluation of target--span linking and its impact on moderation decision-making is deferred to future work, once the pilot corpus is extended.

\textbf{Task description.}
\label{proposed-metho-task-description}
Our approach consists of natively identifying AL spans and corresponding target spans in a text, establishing explicit links between them. 
This structuring differs fundamentally from existing approaches: HX~\cite{Mathew_Saha_Yimam_Biemann_Goyal_Mukherjee_2021} uses token-level annotations only for post-hoc explainability of ALC models, while \citet{pavlopoulos-etal-2021-semeval} is limited to very short toxic spans (approximately 1 word) without target identification. 
Our methodology covers all types of AL, from non-targeted profanity to hate speech, and can capture longer contextual spans, with on average 3 (implicit AL) to 7 (explicit AL) AL tokens per abusive message, essential for characterizing implicit AL.

% To cover all phenomena encountered in industrial context, we consider the following four categories as AL (illustrated in Figure~\ref{tagging_samples}):
% \begin{itemize}
%     \item \textbf{Abusive terms}: vulgarities and insults not directed at a specific entity;
%     \item \textbf{Offensive speech}: targeted AL encompassing personal insults, disrespect, or ironic denigration;    % (commonly defined as offensive or abusive speech in following research~\citet{davidson2017icwsm, Mathew_Saha_Yimam_Biemann_Goyal_Mukherjee_2021});
%     \item \textbf{Threat}: expression of wishes or plans for moral/physical harm against a target;
%     \item \textbf{Hate speech}: incitement to violence, hatred, or discrimination based on protected characteristics (race, religion, gender, etc.).
% \end{itemize}

% Our AL definition stands out by explicitly integrating non-targeted \textbf{Abusive terms}, a category often ignored or conflated with derogatory content in binary approaches. 
% This distinction is crucial for generalization: it avoids false positives on non-abusive intensifier uses. 
% For example, in ``He's a fu*king good player'', \textit{fu*king} is annotated as \textit{Abusive terms} with no link to a target, clearly distinguishing intensifier use from aggressive use.
% For \textbf{Offensive speech}, we follow established definitions of targeted insults and disrespect~\cite{davidson2017icwsm, Mathew_Saha_Yimam_Biemann_Goyal_Mukherjee_2021}, but explicitly extend this category to ironic denigration, a subtle form of targeted disrespect that models rarely detect unless it is explicitly annotated.
% For \textbf{Threat}, we take inspiration from prior work on directed threats~\cite{10.1145/3366424.3385778, vidgen-etal-2021-introducing}, adapting it to operational moderation by isolating expressions of physical or moral harm against a specific target which is a distinction essential for triggering responses that go beyond standard content removal.
% For \textbf{Hate speech}, we align with established definitions~\cite{basile-etal-2019-semeval, caselli-etal-2020-feel}, distinguishing it from \textit{Offensive speech} and \textit{Threat} by the nature of the discourse.

To cover all phenomena encountered in industrial context, we define four AL categories, illustrated in Figure~\ref{tagging_samples}.

\textbf{Abusive terms} are vulgarities and insults not directed at a specific entity.
Our definition stands out by explicitly integrating this non-targeted category, often ignored in binary approaches; it avoids false positives on intensifier uses, as in ``He's a fu*king good player'', where \textit{fu*king} is annotated as \textit{Abusive terms} with no target link.

\textbf{Offensive speech} encompasses targeted personal insults, disrespect, or ironic denigration~\cite{davidson2017icwsm, Mathew_Saha_Yimam_Biemann_Goyal_Mukherjee_2021}.
We explicitly extend this category to ironic denigration, a subtle form of targeted disrespect rarely detected unless annotated.

\textbf{Threat} covers expressions of moral or physical harm against a target~\cite{10.1145/3366424.3385778, vidgen-etal-2021-introducing}, including hypothetical, subtle, and conditional formulations.
In Figure~\ref{tagging_samples}, ``I think, he will kill you'' is annotated as \textit{Threat} even though the harm is reported rather than directly uttered: in an operational setting, such a threat remains informative and relevant to strategic decision-making.
Isolating directed threats is essential for operational moderation, where responses may go beyond standard content removal.

\textbf{Hate speech} denotes incitement to violence, hatred, or discrimination based on protected characteristics (race, religion, gender, etc.)~\cite{basile-etal-2019-semeval, caselli-etal-2020-feel}.
We distinguish it from \textit{Offensive speech} and \textit{Threat} by the nature of the underlying discourse.

% OLD: \paragraph{Experimental scope.}
% OLD: Although our full methodology integrates explicit linking between spans and targets (arrows in Figure~\ref{tagging_samples}), the experiments presented in this article focus on the first step, detecting AL spans, in order to establish a rigorous comparison with ALC approaches. Complete evaluation of target--span linking and its impact on moderation decision-making will be the subject of in-depth evaluation in future work, once the pilot corpus is extended.

\textbf{Corpus creation.}
We built a pilot corpus of 1002 texts from our production moderation pipeline, fed by several public social media platforms combining micro-blogging and thematic discussion forums
\footnote{Our pilot corpus is available for academic collaborations under a confidentiality agreement (NDA). To support reproducibility, we release the experiment code and fine-tuned models at \url{URL}.}
% \footnote{Corpus available for academic collaborations under a confidentiality agreement (NDA), along with code to reproduce the experiments, which will be published upon article publication.}. 
The collection strategy aimed to maximize diversity of AL phenomena while ensuring representativeness of cases encountered in industrial moderation.

No community is targeted by default: the collection objective is to cover every type of abusive discourse, without restricting the sample to any particular community.
Data selection combined flagged keywords with neutral messages from general feeds to balance the distribution, and an automatic pre-filtering step that retained undetected messages specifically to evaluate ALI on ambiguous and implicit cases.

The two professional annotators, one man and one woman, have a background in NLP annotation for sentiment analysis, and the expert annotator adjudicated disagreements to produce the finalized labels~\footnote{The authors actively participated in the annotation as expert annotators.}.
The man annotator has a background in threats annotations, and the woman annotator has a novice for this annotation task.
A formation session was conducted to familiarize the annotators with the ALI task and the annotation protocol, and the annotation conditions took the annotators' well-being into account.
% OLD: Data selection took place in two phases. First, we targeted user profiles and online communities known for at-risk behavior, including archives of moderated thematic forums and public accounts identified as frequent sources of polarized discourse or targets of harassment campaigns. This targeting was complemented by index terminology (flagged keywords and expressions) to identify potentially problematic content, as well as strategic insertion of neutral (non-abusive) messages from general feeds to balance the distribution and avoid prevalence bias. Second, an initial ALC-style automatic filtering step was applied to prioritize samples for annotation, while retaining a subset of messages not detected by this preliminary filter to evaluate ALI capabilities on ambiguous or implicit cases.

The resulting corpus was annotated according to the ALI protocol described above.
It covers detection of AL spans and their associated targets (level 1), as well as classification according to whether the AL is implicit or explicit. 
Among the 1002 finalized texts, 542 contain explicit AL, 123 implicit AL (requiring contextual interpretation to identify abusive character), and 337 are neutral. 
Inter-annotator agreement is computed over three annotators: one expert and the two professional annotators. 
The mean Cohen’s $\kappa$ across pairwise comparisons \cite{cohen1960coefficient} reaches 0.52 for sentence-level AL presence and 0.47 for targets, and drops to 0.38 at the token level.
Although this figure falls below the threshold often considered moderate agreement and therefore injects noise into the ALI training signal, it reflects the intrinsic difficulty of span-boundary and subcategory decisions; comparable span-labeling corpora report similarly constrained agreement despite simpler schemes (e.g., SemEval-2021 toxic spans~\cite{pavlopoulos-etal-2021-semeval}).
We provide more detailed statistics by category in Appendix~\ref{statistique-corpus}.

% !TEX root = acl_latex_en.tex
\section{Experimental setup}

In this section, we describe the models and datasets used to compare the ALI and ALC tasks, as well as training parameters.
% OLD: We note that the experiments presented focus on \textbf{detection of AL spans}, leaving complete evaluation of linking with targets to future work.

\subsection{Language models}
We fine-tuned three models based on transformers commonly used in the AL domain.
\textbf{RoBERTa-large}~\cite{liu2019robertarobustlyoptimizedbert} is used as a generalist reference model, without specific pre-training on AL, allowing us to evaluate the contribution of the ALI task independently of pre-fine-tuning on abusive data.
\textbf{HateBERT}~\cite{caselli2021hatebertretrainingbertabusive} is a BERT variant fine-tuned on a large AL corpus covering various forms of offensive language.
\textbf{fBERT}~\cite{sarkar2021fbertneuraltransformeridentifying} is also a BERT variant, fine-tuned on a corpus of more than 9 million offensive messages, enabling it to capture subtle abusive content.

\textbf{Training parameters.}
For each model, we perform 3 training runs with varying random seeds.
Training is capped at 50 epochs with early stopping (patience of 5 epochs) and a learning rate of $10^{-5}$.

\subsection{Training sets}
We describe below the corpora used to fine-tune the models on the two tasks studied.
We did not retain the corpus of \citet{pavlopoulos-etal-2021-semeval} because annotated spans are mostly isolated lexical units (average length $\approx$ 1 word), incompatible with our goal of capturing contexts for implicit AL.

\textbf{Our proposed corpus.}
In line with the literature on implicit AL detection~\cite{elsherief-etal-2021-latent}, we initially distinguish implicit and explicit classes in annotation, but merge them for ALI/ALC comparison experiments, as ALC cannot distinguish these nuances.
We split the corpus into 80\% training / 20\% validation.

For the ALI task, we use a standard sequential labeling scheme \textit{IO} where label \textit{I} marks lexical units that are part of an AL span, and \textit{O} the rest of the text. For the ALC task, we consider a text abusive if it contains at least one \textit{I} unit. To ensure a fair comparison, we apply the same heuristic in post-processing on ALI predictions (a text is positive if the model predicts at least one AL token), allowing use of the same metrics as ALC.

\textbf{HateXplain corpus.}
Since our corpus is small ($n=1002$), we also use HateXplain ~\cite{Mathew_Saha_Yimam_Biemann_Goyal_Mukherjee_2021}, named here (HX), to validate the ALI approach at larger scale ($\approx$17k sentences).
This corpus contains token-level annotations (rationales) justifying the classification decision, which we reinterpret as AL spans for the ALI task.

However, HX exhibits heterogeneity in annotations: rationales vary significantly between annotators.
For example, two annotators may agree on the \textit{hateful} label while highlighting entirely different spans as the justification.
For the ALI task, we randomly extract a single justification per text to simulate standard sequence labeling annotation.
Furthermore, some texts annotated as abusive have no rationale; in those cases, we treat them as negative for ALI training (no detectable span).

We use the standard HX split (15k train / 2k val), with 59.4\% abusive texts.
The same preprocessing and post-processing are applied to HX and our corpus to ensure methodological uniformity.

\textbf{Target-based offensive corpus.}
We also use the corpus provided by \citet{zampieri-etal-2023-target} to evaluate the generalization of ALI vs ALC.
The corpus combines sentence-level harmfulness labels with token-level annotations of offensive argument spans and their associated targets.
At the token level, each unit is labeled as part of an offensive argument, as a target entity, or as outside any span.
We apply the same preprocessing and post-processing as for previously described corpora to ensure methodological uniformity.
Unfortunately, only the training set is publicly available; we split it into 80\% training and 20\% validation sets for our experiments.

\textbf{Training combination.}
We study the impact of merging the two corpora HX and TBO (each merged independently) with our pilot corpus for training.
We anticipate that ALI models will be more sensitive to these training combinations than ALC models, due to the greater specificity of the token-level annotation objective (risk of inconsistency in spans between the two sources).

We also experimented with automatic annotation via generative LLMs\footnote{Mistral-Nemo, LLaMa-3.3-70B, Qwen2.5-32B; precision $<$ 70\% on an internal validation sample.} but results were insufficient for direct integration, probably due to the complexity of implicit/explicit disambiguation; a dedicated study is left to future work.

\subsection{Evaluation sets and metrics}
We evaluate our models in an out-of-domain configuration on HatEval, IHC, JL-Hate, and {TR}u{ST}, to measure generalization of ALI vs ALC.
We do not evaluate on the corpus provided by \citet{zampieri-etal-2023-target} because the test set is not publicly available.
However, we provide sequence labeling evaluation on the JL-Hate and {TR}u{ST} corpora because they contain token-level annotations of AL spans.

\textbf{HatEval corpus.}
We use the test set of \citet{basile-etal-2019-semeval} (approximately 2900 sentences annotated for hate speech detection).
For comparison with ALI, we binarize predictions (hateful = abusive, non-hateful = neutral).
This dataset is standard for evaluating generalization~\cite{10.1145/3331184.3331262, caselli-etal-2020-feel, YUAN2025101690}, notably because it contains counter-responses to hate speech (counter-discourse) that sometimes use abusive language, a textbook case for testing model robustness.
We do not use this corpus for span-level analysis, because it has no token-level annotations.

\textbf{IHC corpus.}
This corpus~\cite{elsherief-etal-2021-latent} is annotated in three classes: \textit{non-hateful}, \textit{explicit hateful}, and \textit{implicit hateful}.
We randomly sample 1000 texts per class and merge explicit/implicit classes into a single \textit{abusive} class for ALI/ALC comparison.
IHC enables fine-grained analysis of implicit AL detection, a key phenomenon for validating the interest of ALI.
We do not use this corpus for span-level analysis, because it has no token-level annotations.

\textbf{JL-Hate corpus.}
JL-Hate~\cite{buyukdemirci-etal-2024-jl} combines three sentence-level classes (\textit{non-abusive}, \textit{offensive}, \textit{hateful}) with token-level annotations of AL and target spans.
We use the full English subset (765 sentences) as an out-of-domain evaluation set: models are never trained on this corpus.
Sentence-level evaluation follows the same protocol as HatEval and IHC (merged \textit{abusive} class; ALI post-processed for macro-F1).
We additionally use this corpus for span-level evaluation based on AL and target spans.

\textbf{{TR}u{ST} corpus.}
{TR}u{ST}~\cite{atil2026trust} unifies prior social-media and LLM-generated resources under a common scheme for binary toxicity, targeted social groups, and toxic spans.
We use its human-annotated test split (982 texts).
Sentence-level evaluation follows the same protocol as the other corpora, with binary toxicity mapped to our abusive / non-abusive labels and \textit{unsure} texts excluded.
Span-level evaluation is restricted to AL spans because targets are annotated as categorical groups rather than spans.

\textbf{Evaluation metrics.}
For sentence level evaluation, we report \textit{macro-F1}, standard in abusive language detection~\cite{basile-etal-2019-semeval, caselli-etal-2020-feel}.
For ALI, we derive a sentence-level label by binarizing token predictions: a text is \textit{abusive} if at least one token is predicted as AL.
This post-processing yields the same sentence-level macro-F1 for ALI and ALC on HatEval, IHC, JL-Hate, and {TR}u{ST}.
For each system, we select the run with the median macro-F1 among three random initializations, then compare ALI and ALC predictions on each test set with a significance test~\cite{gillick-cox-1989} at a 5\% level.

Concerning span-level evaluation, we report F1-exact, F1-partial, and F1-token metrics for both AL and Target spans identification:
F1-exact requires identical predicted and gold span boundaries;
F1-partial credits a prediction when it overlaps the gold span;
F1-token is the binary macro-F1 over tokens..
We report median scores and standard deviation over three training runs.
Concerning significance tests, we apply similar tests as for sentence-level evaluation: the significance test at 5\% on the median run.

% !TEX root = acl_latex_en.tex
\section{Results}

\subsection{Span level obtained results}
\label{section:span-level-obtained-results}
Table~\ref{table:span-level-obtained-results} compares standard ALI and \textit{with target} models on JL-Hate and {TR}u{ST} (\textit{Ours}, \textit{TBO}).
An exhaustive analysis of all training settings (including HX and other training combinations) is provided in Appendix~\ref{annex:span-level-obtained-results}.
\begin{table*}[t]
    \begin{center}
    \resizebox{\textwidth}{!}{%
    \begin{tabular}{llccccccccc}
    \toprule
    \multirow{2}{*}{Models} & \multirow{2}{*}{Training sets} & \multicolumn{3}{c}{JL-Hate AL} & \multicolumn{3}{c}{JL-Hate Target} & \multicolumn{3}{c}{{TR}u{ST} AL} \\
    \cmidrule(lr){3-5} \cmidrule(lr){6-8} \cmidrule(lr){9-11}
     & & F1-exact & F1-partial & F1-token & F1-exact & F1-partial & F1-token & F1-exact & F1-partial & F1-token \\
    \midrule
    \multirow{4}{*}{fBERT} & Ours & \apm{18.20}{0.40} & \apm{\textbf{51.85}}{0.34}$^\dagger$ & \apm{67.72}{0.55} & -- & -- & -- & \apm{13.62}{1.04} & \apm{61.23}{0.80} & \apm{48.53}{0.36} \\
     & Ours (with target) & \apm{17.49}{0.90} & \apm{51.75}{0.64} & \apm{67.66}{0.43} & \apm{\textbf{24.08}}{0.92}$^\dagger$ & \apm{\textbf{38.86}}{0.37}$^\dagger$ & \apm{\textbf{66.17}}{0.64}$^\dagger$ & \apm{13.70}{1.25} & \apm{59.45}{0.71} & \apm{48.99}{0.89} \\
     & TBO & \apm{\textbf{26.52}}{0.35} & \apm{51.61}{0.88} & \apm{\textbf{71.40}}{0.55} & -- & -- & -- & \apm{27.50}{0.88} & \apm{60.09}{0.57} & \apm{\textbf{49.31}}{0.78}$^\dagger$ \\
     & TBO (with target) & \apm{25.93}{0.48} & \apm{50.85}{0.44} & \apm{71.03}{0.48} & \apm{18.49}{1.13} & \apm{29.82}{0.78} & \apm{61.05}{0.15} & \apm{\textbf{27.58}}{1.66} & \apm{\textbf{62.15}}{0.70} & \apm{48.25}{0.30} \\
    % & Ours + TBO & \apm{22.28}{1.15} & \apm{39.95}{0.93} & \apm{65.86}{0.27} & -- & -- & -- & \apm{22.20}{2.15} & \apm{41.73}{2.68} & \apm{41.15}{0.41} \\
    % & Ours + TBO (with target) & \apm{23.95}{0.53} & \apm{45.45}{1.61} & \apm{67.87}{0.71} & \apm{21.78}{2.36} & \apm{34.42}{1.78} & \apm{63.08}{0.84} & \apm{24.43}{1.45} & \apm{42.75}{1.95} & \apm{41.20}{0.39} \\
    \addlinespace
    \multirow{4}{*}{HateBERT} & Ours & \apm{19.07}{0.83} & \apm{54.09}{1.08} & \apm{67.98}{0.82} & -- & -- & -- & \apm{12.81}{0.32} & \apm{62.74}{0.25} & \apm{48.01}{0.21} \\
     & Ours (with target) & \apm{19.35}{1.02} & \apm{\textbf{54.63}}{0.59}$^\dagger$ & \apm{68.41}{0.17} & \apm{\textbf{24.06}}{1.29} & \apm{\textbf{37.74}}{0.51} & \apm{\textbf{65.93}}{0.10} & \apm{16.33}{1.70} & \apm{\textbf{63.27}}{1.04} & \apm{\textbf{48.33}}{0.18} \\
     & TBO & \apm{\textbf{27.06}}{0.49}$^\dagger$ & \apm{52.61}{1.34} & \apm{\textbf{71.52}}{0.41}$^\dagger$ & -- & -- & -- & \apm{\textbf{28.18}}{1.49} & \apm{60.91}{0.37} & \apm{47.07}{0.64} \\
     & TBO (with target) & \apm{25.69}{1.62} & \apm{50.62}{0.93} & \apm{71.48}{0.42} & \apm{21.06}{2.16} & \apm{32.82}{1.62} & \apm{62.88}{0.89} & \apm{27.91}{0.52} & \apm{61.22}{0.38} & \apm{47.62}{0.24} \\
    % & Ours + TBO & \apm{21.29}{1.47} & \apm{40.35}{2.02} & \apm{65.61}{1.07} & -- & -- & -- & \apm{24.44}{1.02} & \apm{44.37}{0.33} & \apm{41.48}{0.08} \\
    % & Ours + TBO (with target) & \apm{19.40}{0.73} & \apm{37.26}{2.01} & \apm{63.89}{1.09} & \apm{21.20}{0.32} & \apm{35.38}{1.02} & \apm{63.83}{0.42} & \apm{21.37}{2.73} & \apm{39.46}{3.28} & \apm{41.07}{0.50} \\
    \addlinespace
    \multirow{4}{*}{RoBERTa} & Ours & \apm{19.22}{2.79} & \apm{53.95}{1.28} & \apm{69.04}{1.37} & -- & -- & -- & \apm{15.19}{0.77} & \apm{64.75}{0.61} & \apm{50.41}{0.66} \\
     & Ours (with target) & \apm{22.53}{0.68} & \apm{\textbf{56.67}}{1.05}$^\dagger$ & \apm{68.16}{0.48} & \apm{\textbf{31.13}}{0.21} & \apm{\textbf{43.35}}{0.78} & \apm{\textbf{70.45}}{0.58} & \apm{15.89}{0.40} & \apm{\textbf{66.00}}{0.35} & \apm{\textbf{51.17}}{0.49} \\
     & TBO & \apm{27.67}{0.60} & \apm{54.10}{0.34} & \apm{71.71}{0.31} & -- & -- & -- & \apm{27.96}{1.52} & \apm{60.82}{1.24} & \apm{48.79}{0.22} \\
     & TBO (with target) & \apm{\textbf{27.89}}{0.87} & \apm{53.52}{0.36} & \apm{\textbf{72.55}}{0.18} & \apm{30.61}{0.81} & \apm{43.29}{1.11} & \apm{68.03}{0.43} & \apm{\textbf{28.34}}{1.00} & \apm{60.59}{0.92} & \apm{48.29}{0.57} \\
    % & Ours + TBO & \apm{18.90}{0.27} & \apm{35.20}{2.76} & \apm{64.52}{0.66} & -- & -- & -- & \apm{19.63}{1.17} & \apm{34.58}{4.39} & \apm{40.78}{0.62} \\
    % & Ours + TBO (with target) & \apm{20.05}{2.40} & \apm{35.36}{3.54} & \apm{64.15}{1.59} & \apm{\textbf{32.30}}{0.78} & \apm{\textbf{43.48}}{0.62} & \apm{69.40}{0.65} & \apm{18.95}{0.63} & \apm{37.19}{2.16} & \apm{40.62}{0.35} \\
    \bottomrule
    \end{tabular}%
    }
    \caption{Median F1$\pm$std (\%, 3 runs) for AL/Target spans on JL-Hate and AL spans on {TR}u{ST}.
    \textbf{Bold} text marks the best score for each F1 type within a backbone, and the $^\dagger$ symbol is used to indicate that the corresponding score is significantly better than the other scores of the same backbone.}
    \label{table:span-level-obtained-results}
    \label{table:trust-ood-spans}
    \end{center}
\end{table*}

\textbf{AL span detection.}
Across backbones, TBO-based training yields the highest F1-exact and F1-token, while training on our corpus with target supervision can lead on F1-partial (Table~\ref{table:span-level-obtained-results}).
F1-exact stays well below F1-partial and F1-token on both JL-Hate and {TR}u{ST}, remaining below $30$\%.
The target head does not help uniformly: for RoBERTa trained on our corpus it raises F1-exact and F1-partial ($22.53$ vs $19.22$; $56.67$ vs $53.95$), while F1-token remains slightly higher without it ($69.04$ vs $68.16$).
Combining our corpus with TBO brings no further gain, probably because the two annotation schemes are too dissimilar.
On {TR}u{ST} test set, we observe that F1-token falls to $47.07$--$51.17$\% from an F1-partial of $59.45$--$66.00$\%, because some texts are annotated as entirely toxic: F1-partial credits any overlap, while F1-token penalizes every unmarked gold token.

\textbf{Target detection.}
Target spans are annotated only on JL-Hate.
{TR}u{ST} labels targets as categorical groups rather than spans, so no target-span score can be reported.
With target supervision, \textit{Ours (with target)} is the strongest configuration for fBERT and significantly better than the other with-target rows of that backbone ($^\dagger$).
For HateBERT and RoBERTa, gaps against the other with-target trainings are not significant under this best-vs-all criterion.
RoBERTa still reaches the highest target scores, at $31.13$ F1-exact, $43.35$ F1-partial, and $70.45$ F1-token on our corpus.
Despite the small size of our pilot corpus, these results suggest that our annotation scheme provides relevant supervision for joint AL and target span detection.

Overall, accurately detecting both AL and target spans remains highly challenging, particularly when exact boundaries are required.

\subsection{ALI vs ALC}
We compare ALI and ALC on cross-domain generalization and robustness on implicit AL; sentence-level ALI labels are derived by binarizing token predictions (a text is abusive if at least one token is predicted as AL).
Table~\ref{median-macro-ali-vs-classif} focuses on the three single-source trainings (\textit{Ours}, \textit{HX}, \textit{TBO}) to isolate the ALI vs ALC contrast on HatEval, IHC, JL-Hate, and the {TR}u{ST}. 
The full resutl grid is reported in Appendix~\ref{annex:ali-vs-alc-full}.
\begin{table*}[t]
    \begin{center}
    \resizebox{\textwidth}{!}{%
    \begin{tabular}{@{}l@{~}lc@{~}c@{~~}c@{~}c@{~~}c@{~}c@{~~}c@{~}c@{}}
    \toprule
    Models & Training sets & \multicolumn{2}{c}{HatEval} & \multicolumn{2}{c}{IHC} & \multicolumn{2}{c}{JL-Hate} & \multicolumn{2}{c}{{TR}u{ST}} \\
    \cmidrule(lr){3-4} \cmidrule(lr){5-6} \cmidrule(lr){7-8} \cmidrule(lr){9-10}
     & & ALI & ALC & ALI & ALC & ALI & ALC & ALI & ALC \\
    \midrule
    \multicolumn{2}{l}{\citet{10.1145/3331184.3331262}} & -- & 54.1 & -- & -- & -- & -- & -- & -- \\
    \multicolumn{2}{l}{\citet{caselli-etal-2020-feel}} & -- & 59.1 & -- & -- & -- & -- & -- & -- \\
    \multicolumn{2}{l}{\citet{bose-etal-2021-unsupervised}} & -- & 57.8 & -- & -- & -- & -- & -- & -- \\
    \multicolumn{2}{l}{\citet{YUAN2025101690}} & -- & 64.5 & -- & -- & -- & -- & -- & -- \\
    \midrule
    \multirow{3}{*}{fBERT} & Ours & \apm{54.64}{0.80}$^*$ & \apm{53.28}{0.11} & \apm{63.21}{0.83} & \apm{62.44}{0.26} & \apm{69.82}{0.81} & \apm{69.01}{0.67} & \apm{\textbf{71.97}}{0.43} & \apm{\textbf{74.22}}{0.07}$^\dagger$ \\
     & HX & \apm{\textbf{67.15}}{0.08}$^\dagger$ & \apm{\textbf{67.87}}{0.30}$^\dagger$ & \apm{\textbf{68.00}}{0.21}$^{*\dagger}$ & \apm{\textbf{65.70}}{0.86}$^\dagger$ & \apm{67.44}{0.57} & \apm{\textbf{71.14}}{3.98}$^*$ & \apm{68.34}{0.53} & \apm{70.62}{1.98} \\
     & TBO & \apm{50.06}{0.38}$^*$ & \apm{48.44}{1.64} & \apm{59.90}{3.21}$^*$ & \apm{42.55}{1.95} & \apm{\textbf{69.88}}{1.34}$^*$ & \apm{43.62}{0.55} & \apm{68.28}{0.21}$^*$ & \apm{43.35}{1.03} \\
     \addlinespace
    \multirow{3}{*}{HateBERT} & Ours & \apm{55.84}{0.60}$^*$ & \apm{43.29}{5.26} & \apm{60.66}{0.75}$^*$ & \apm{48.83}{5.00} & \apm{70.26}{0.96}$^*$ & \apm{47.55}{9.28} & \apm{\textbf{73.83}}{0.53}$^*$ & \apm{58.90}{12.30} \\
     & HX & \apm{\textbf{61.80}}{1.07}$^\dagger$ & \apm{\textbf{67.17}}{0.26}$^{*\dagger}$ & \apm{\textbf{68.72}}{0.31}$^{*\dagger}$ & \apm{\textbf{65.59}}{0.16}$^\dagger$ & \apm{69.17}{1.97} & \apm{\textbf{68.68}}{0.84}$^\dagger$ & \apm{67.09}{0.63} & \apm{\textbf{70.70}}{0.65}$^\dagger$ \\
     & TBO & \apm{49.86}{1.18}$^*$ & \apm{39.05}{0.40} & \apm{59.46}{0.83}$^*$ & \apm{27.41}{0.11} & \apm{\textbf{73.07}}{0.14}$^*$ & \apm{41.12}{0.56} & \apm{71.21}{0.21}$^*$ & \apm{36.29}{0.28} \\
    \addlinespace
    \multirow{3}{*}{RoBERTa} & Ours & \apm{52.20}{0.68}$^*$ & \apm{47.20}{0.89} & \apm{64.38}{0.47} & \apm{63.16}{13.9} & \apm{71.05}{1.11} & \apm{\textbf{72.29}}{15.7}$^\dagger$ & \apm{\textbf{75.53}}{0.34} & \apm{\textbf{71.57}}{19.04} \\
     & HX & \apm{\textbf{65.58}}{0.86}$^\dagger$ & \apm{\textbf{65.06}}{1.48}$^\dagger$ & \apm{\textbf{69.98}}{0.58}$^{*\dagger}$ & \apm{\textbf{66.82}}{1.08}$^\dagger$ & \apm{\textbf{71.51}}{1.39}$^*$ & \apm{65.50}{2.20} & \apm{71.42}{0.94} & \apm{71.05}{1.16} \\
     & TBO & \apm{49.62}{1.23}$^*$ & \apm{43.55}{1.05} & \apm{61.13}{0.32}$^*$ & \apm{39.64}{2.22} & \apm{63.23}{1.88}$^*$ & \apm{36.56}{0.13} & \apm{72.92}{0.03}$^*$ & \apm{40.63}{1.53} \\
    \midrule
    \multicolumn{2}{l}{Average per task} & 56.31 & 52.77 & 63.94 & 53.57 & 69.49 & 57.27 & 71.18 & 59.70 \\
    \bottomrule
    \end{tabular}%
    }

    \caption{Median macro-F1$\pm$std (\%, 3 runs) of ALI and ALC trained on single sources.
    \textit{Average per task} is the mean of the nine medians (published HatEval scores excluded).
    \textbf{Bold} text marks the best score for each approach within a model and test set; the $^*$ symbol is used to indicate that the corresponding score is significantly better in the ALI--ALC comparison for the same training setting, and the $^\dagger$ symbol is used to indicate that it is significantly better than the other training settings of the same model and approach.}
    \label{median-macro-ali-vs-classif}
    \label{table:trust-ood}
    \end{center}
\end{table*}

\textbf{Average ALI vs ALC trend.}
Although ALI natively outputs token-level labels and is evaluated here via post-processing for comparability with ALC, it matches or surpasses ALC across test sets (Table~\ref{median-macro-ali-vs-classif}, \textit{Average per task} row).
The largest gaps appear on JL-Hate ($69.49$\% vs $57.27$\%; $+12.2$) and {TR}u{ST} ($71.18$\% vs $59.70$\%; $+11.5$), then IHC ($63.94$\% vs $53.57$\%; $+10.4$), while HatEval remains closer ($56.31$\% vs $52.77$\%; $+3.5$), consistent with the sequence-labeling design on span-rich corpora.
Under RoBERTa + HX, ALI remains competitive in this ``degraded'' sentence-level setup ($65.58$\% vs $65.06$\% on HatEval; $69.98$\% vs $66.82$\% on IHC; $71.51$\% vs $65.50$\% on JL-Hate), confirming that span-level detection does not degrade global detection while adding interpretability.
On HatEval, RoBERTa ALI reaches $65.58$\% macro-F1, surpassing the $64.5$\% previously reported by \citet{YUAN2025101690}, despite providing localized outputs.
The best ALC score on this set is fBERT with HX training ($67.87$\% macro-F1).

\textbf{Robustness under TBO and limited data.}
ALC collapses under TBO training on several configurations (e.g., HateBERT on IHC: $27.41$\%; JL-Hate: $41.12$\%; {TR}u{ST}: $36.29$\%), whereas ALI remains stable ($59.46$\%, $73.07$\%, and $71.21$\%, respectively).
This gap likely stems from a distributional mismatch: TBO annotates targeted offensive spans, whereas IHC and JL-Hate evaluate broader or differently structured AL phenomena.
ALC appears to overfit TBO's global harmfulness decision, whereas ALI retains transferable local cues even when the training domain differs.
Training on our corpus alone ($n=800$) yields lower HatEval scores ($52$--$56$\% macro-F1 for ALI), as expected given size and class imbalance, but ALI suffers less from data restriction than ALC in several cases (e.g., HateBERT on IHC: $60.66$\% ALI vs $48.83$\% ALC; RoBERTa ALC standard deviation up to $13.9$ on IHC).

\textbf{Backbone differences and training combinations.}
HX is the strongest single-source training for HatEval and IHC across backbones, while TBO remains competitive for ALI on JL-Hate (HateBERT: $73.07$\%).
We hypothesize that RoBERTa-large's stronger results reflect its larger number of trainable parameters relative to HateBERT and fBERT.
Effects of merging corpora (\textit{Ours + HX}, \textit{Ours + TBO}, \textit{TBO + HX}, \textit{Ours + TBO + HX}) are reported in Appendix~\ref{annex:ali-vs-alc-full}: combinations with TBO often degrade ALI on IHC and HatEval, whereas combining our pilot corpus with HX can further improve IHC and JL-Hate.
Appendix~\ref{annex:abusive_language_implicit_explicit} further reports implicit and explicit AL detection on IHC; under settings without TBO, ALI outperforms ALC on implicit AL (+$3.6$ macro-F1; $61.42$\% vs $57.86$\%) without trading off explicit AL detection (+$2.1$).

% !TEX root = acl_latex_en.tex
%%%%%%%%%%%%%%%%%%%%%%%%%%%%%%%%%%%%%%%%%%%
% QUALITATIVE ANALYSIS                    %
%%%%%%%%%%%%%%%%%%%%%%%%%%%%%%%%%%%%%%%%%%%

\section{Qualitative analysis}
\label{annex:analyse-quali}
In this section, we conduct a qualitative analysis by examining predictions on a few samples from the IHC corpus.
We base our analysis on the RoBERTa model, as it obtained the best macro-F1 scores among training sets and tasks used on the IHC test set. 
Furthermore, the RoBERTa model has no pre-training on AL, which makes it useful for conducting this analysis.

\begin{figure*}[t]
\begin{center}
\includegraphics[width=.95\linewidth]{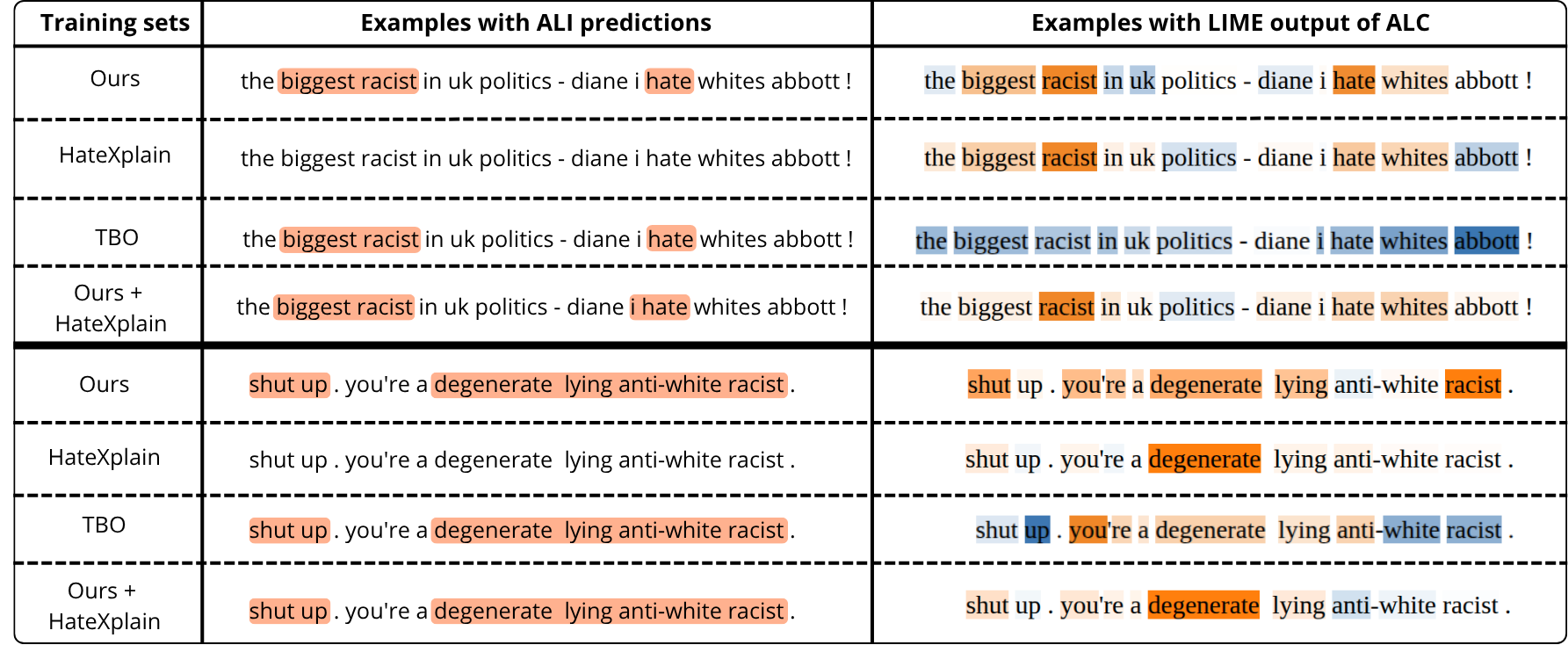}
\caption{Example outputs for RoBERTa ALI and ALC models trained on different corpora. For ALC models, LIME explanations are shown: words highlighted in orange tend to predict AL, and those highlighted in blue the opposite (normal discourse). The more a word is highlighted, the more impact it has on the class predicted by the ALC model.
For ALI models, ALI predictions are highlighted.
Both examples shown are from the IHC test set.
}
\label{ali_vs_alc_lime}
\end{center}
\end{figure*}

Unlike ALC, ALI natively localizes AL in text and therefore does not rely on post-hoc explanation methods, making predictions more directly interpretable for non-expert analysts.
For comparison, we apply LIME~\cite{lime} to ALC models to highlight the words most influential for the classification decision, and contrast these post-hoc explanations with ALI span predictions.
Figure~\ref{ali_vs_alc_lime} illustrates this comparison on two examples from the IHC test set.

\textbf{Labeling divergence and its effect on explanations.}
On the first sample (top of Figure~\ref{ali_vs_alc_lime}), we note that the verb ``\textit{hate}'' is ignored by the ALI model trained on HX, unlike the other models. This is probably due to the low proportion of annotation of this term in HX data: only 24\% of lexical units ``\textit{hate}'' are considered rationales in the training set, vs 41\% in our training set.
Furthermore, we observe the impact of labeling divergences between the two training sets: on the LIME output, we note that ``\textit{hate}'' contributes less to class prediction when the model is trained on the HX training set.
When models are trained on TBO~\cite{zampieri-etal-2023-target}, we observe a pronounced shift in focus between ALI and ALC: ALI localizes offensive arguments in context, whereas ALC explanations remain concentrated by focusing on the wrong interpretation of the text.
These observations suggest that span-level ALI annotation, guided by a consistent taxonomy, can mitigate lexicon-driven biases inherited from training corpora and support more reliable AL localization when several abusive phenomena co-occur in the same text.

The second sample further illustrates cross-corpus annotation divergence: ``\textit{shut up}'' is labeled AL in fewer than 13\% of HX training instances, versus systematically in our corpus, visibly shifting the LIME explanations. With TBO-trained models, a complementary contrast emerges: ALI marks offensive argument spans, whereas ALC concentrates on target mentions.
% OLD: The second sample reinforces this analysis.
% OLD: In our corpus, ``\textit{shut up}'' is systematically annotated as AL, as it may correspond to \textit{Offensive speech}, \textit{Threat}, or \textit{Hate speech} depending on context; in HX, the same expression is labeled as AL in fewer than 13\% of cases, compared with about 40\% for ``\textit{degenerate}''.
% OLD: Accordingly, ALC explanations tend to highlight isolated words, whereas ALI trained on our corpus localizes full multi-word expressions (e.g., ``\textit{shut up}'') and longer phrases (e.g., ``\textit{degenerate lying anti-white racist}'').
% OLD: The same contrast appears with TBO-trained models: ALI marks offensive argument spans in context, while ALC explanations concentrate on target mentions.

\textbf{Distinction of AL subcategories.}
With our training set, most errors arise because models are not trained to distinguish AL subcategories, even though our annotation scheme defines four types: \textit{Abusive terms}, \textit{Offensive speech}, \textit{Threat}, and \textit{Hate speech} (section~\ref{proposed-metho-task-description}).
In ``\textit{no woman has the right to kill an unborn baby! it's not her body it's the baby's body she's choosing to kill. open your eyes !!!}'', the model correctly identifies ``\textit{choosing to kill}'' as AL, but without subcategory supervision it cannot assign it to \textit{Threat}.
This is nonetheless a minor error in practice, since the span itself is appropriately localized.

\textbf{Missing context.}
Our approach also struggles when abusive meaning relies on external cultural knowledge.
In the example ``\textit{maga white cookouts would be nice.}'', annotated as implicit hate in IHC, the model predicts no AL span.
Without familiarity with ``\textit{maga}'' (a U.S. political slogan) and the implicit connotations of ``\textit{white cookouts}'', the message appears benign at the lexical level.
Such errors suggest that implicit AL detection depends on world knowledge that span-level modeling alone cannot fully capture, and may require broader training coverage of implicit phenomena or access to additional contextual information.

% !TEX root = acl_latex_en.tex

\section{Conclusion and future work}

\textbf{Conclusion.}
% In this paper, we compared abusive language identification (ALI), a sequence labeling task, with abusive language classification (ALC) across three transformer encoders and multiple training corpora, evaluating cross-domain generalization on HatEval, IHC, JL-Hate, and {TR}u{ST} under a conservative sentence-level protocol for ALI predictions.
% At the span level, ALI locates AL segments reasonably well on overlap-based metrics but struggles with exact boundaries (F1-exact below $30$\% on JL-Hate and {TR}u{ST}), and target spans remain even harder to recover (best F1-token of $70.45$\% with RoBERTa), confirming that fine-grained AL and target detection is still a challenging task.
% At the sentence level, ALI nonetheless matches or outperforms ALC on average across all test sets, with the largest gains on JL-Hate ($+12.2$ macro-F1) and {TR}u{ST} ($+11.5$), while remaining far more stable when trained on TBO.
% Our qualitative analyses highlight the difficulties faced by ALI models trained on our small pilot corpus, and underscore the need for annotation according to the hierarchical AL taxonomy.
In this paper, we compared abusive language identification (ALI), a sequence labeling task, with abusive language classification (ALC) across three transformer encoders and multiple training corpora, evaluating cross-domain generalization on HatEval, IHC, JL-Hate, and {TR}u{ST} under a conservative sentence-level protocol for ALI predictions.
We also provided span level analysis, where ALI recovers AL and target segments on overlap-based metrics but struggles with exact boundaries (F1-exact below $30$\% for AL spans and around $31$\% for target spans), showing that fine-grained identification remains challenging.
Conserning sentence level analysis, ALI matches or outperforms ALC on average on every test set despite a conservative binarization of its token predictions, with the largest gains on JL-Hate ($+12.2$ macro-F1) and {TR}u{ST} ($+11.5$) and a narrower margin on HatEval ($+3.5$).
ALI is also more robust: when trained on TBO, ALC collapses on several test sets (e.g., HateBERT on IHC: $27.41$\%) while ALI remains stable ($59.46$\%), and ALI suffers less from the limited size of our pilot corpus.
Finally, ALI improves implicit AL detection over ALC ($+3.6$ macro-F1 on IHC) without trading off explicit AL detection, while additionally providing localized, interpretable outputs.
Our qualitative analyses highlight the difficulties faced by ALI models trained on our small pilot corpus, and underscore the need for annotation according to the hierarchical AL taxonomy.

\textbf{Future work.} 
Three priority directions emerge: 
(1) enriching existing corpora with complete ALI annotations to build structured benchmarks; 
(2) integrating target--AL linking and exploring multi-task architectures exploiting the hierarchical taxonomy; 
(3) systematic bias evaluation, as ALI enables fine-grained error analysis at the token level unlike the ALC task.

\section*{Limitations}
Annotating a corpus using our proposed granularity remains a difficult and costly task: token-level agreement is limited (mean pairwise Cohen's $\kappa=0.38$), reflecting the complexity of span-boundary decisions.
Our study nevertheless demonstrates the feasibility of ALI with approximately 1000 samples (English-only), sufficient to validate the approach, although a larger scale (10k+ samples) would be required to reach the operational robustness expected in production.
Our qualitative analysis shows that confusion between subcategories (notably \textit{Threat} and \textit{Hate speech}) reflects the intrinsic difficulty of fine taxonomy on short spans, and partly explains lower scores on HatEval where this distinction is semantically critical.
The out-of-domain corpora use heterogeneous definitions and annotation schemes, which limits the direct comparability of results across datasets.
Target--AL linking remains preliminary because our experiments identify target spans without explicitly associating each target with its corresponding AL span.

\section*{Ethical considerations}

Handling data containing abusive language raises important ethical concerns. An explicit warning about the sensitive nature of the content was provided to annotators from the start of the process. Regular rest periods and emotional support were made available, and they were free to discontinue participation at any time. We are aware that the datasets used may contain biases related to annotator choices or data sources, and we encourage future work to mitigate them.

\bibliography{custom}

\appendix
% !TEX root = acl_latex_en.tex

\section{Statistics of our pilot corpus}
\label{statistique-corpus}

\begin{table}[h]
    \centering
    \small
    \begin{tabular}{@{~}l@{~}c@{~}@{~}c@{~}@{~}c@{~}}
        \toprule
         & Neutral & Implicit AL & Explicit AL \\
         \midrule
         \textbf{Sentences (n)} & 337 & 123 & 542 \\
         \midrule
         \multicolumn{4}{l}{\textit{Statistics on sentences}} \\
         Max. length (tokens) & 101 & 80 & 90 \\
         Avg. length (tokens) & 26.64 & 22.73 & 25.32 \\
         \midrule
         \multicolumn{4}{l}{\textit{Statistics on AL tokens}} \\
         Max. AL tokens per sentence & $\emptyset$ & 29 & 61 \\
         Avg. AL tokens per sentence & $\emptyset$ & 3.08 & 7.24 \\
         \bottomrule
    \end{tabular}
    \caption{Pilot corpus statistics by class. \textit{AL tokens} are tokens tagged as part of an AL span.}
    \label{tab:dataset-specification}
\end{table}

\section{Implicit and explicit abusive language detection}
\label{annex:abusive_language_implicit_explicit}

In Table~\ref{median-macro-implicit-explicit-hate}, we present results
obtained by focusing on implicit and explicit abuse in IHC.
As part of a more in-depth analysis, we evaluate our approach
by focusing on implicit or explicit abusive discourse: we test on
an IHC subset containing, on one hand, \textit{implicit} and
\textit{neutral} labels (column \textit{Implicit AL} in Table~\ref{median-macro-implicit-explicit-hate}),
and on the other hand \textit{explicit} and \textit{neutral} labels (column
\textit{Explicit AL} in Table~\ref{median-macro-implicit-explicit-hate}).

We observe on average (see the \textit{Average per task} row in
Table~\ref{median-macro-implicit-explicit-hate}) that the ALI task surpasses the ALC task
for implicit and explicit AL detection: ALI reaches 52.20\%
and 65.22\% average macro-F1 scores for implicit and
explicit AL detection respectively, vs 52.04\% and 63.08\% average macro-F1 obtained
with the ALC task.
These results continue to argue in favor of the effectiveness of our proposed approach.
The performance difference can be explained by the fact that with the ALI task, a model
learns to identify abusive segments in a text and therefore pays more attention
to abusive discourse, which helps it distinguish implicit content from neutral content.
It is important to note that this improvement in implicit AL
detection does not come at the expense of explicit AL, as can be
seen in Table~\ref{median-macro-implicit-explicit-hate}.

\begin{table*}[ht]
\begin{center}
\begin{tabular}{llcccc}
\toprule
Models & Training sets & \multicolumn{2}{c}{Implicit AL} & \multicolumn{2}{c}{Explicit AL} \\
\cmidrule(lr){3-4} \cmidrule(lr){5-6}
 & & ALI & ALC & ALI & ALC \\
\midrule
\multirow{7}{*}{fBERT} & Ours & \apm{57.51}{0.72} & \apm{57.33}{0.60} & \apm{72.06}{0.61}$^*$ & \apm{70.56}{0.28} \\
 & HX & \apm{62.41}{0.30}$^*$ & \apm{58.02}{1.06} & \apm{\textbf{80.35}}{0.43} & \apm{\textbf{78.95}}{0.20} \\
 & TBO & \apm{54.02}{0.06} & \apm{49.34}{0.72} & \apm{65.87}{0.65}$^*$ & \apm{46.63}{1.68} \\
 & Ours + HX & \apm{\textbf{64.50}}{1.19}$^*$ & \apm{\textbf{58.12}}{0.70} & \apm{79.32}{0.17} & \apm{77.94}{0.18} \\
 & Ours + TBO & \apm{42.31}{2.51} & \apm{48.97}{0.72} & \apm{60.37}{2.48}$^*$ & \apm{48.42}{1.50} \\
 & TBO + HX & \apm{42.33}{1.43} & \apm{49.97}{0.23} & \apm{49.59}{1.43} & \apm{66.80}{0.72}$^*$ \\
 & Ours + TBO + HX & \apm{38.54}{1.30} & \apm{51.70}{1.48}$^*$ & \apm{43.34}{1.91} & \apm{69.58}{0.60}$^*$ \\
\addlinespace
\multirow{7}{*}{HateBERT} & Ours & \apm{55.25}{1.03}$^*$ & \apm{45.69}{5.60} & \apm{70.90}{0.20}$^*$ & \apm{50.78}{8.20} \\
 & HX & \apm{63.05}{0.24}$^*$ & \apm{\textbf{60.38}}{0.34} & \apm{77.64}{0.21} & \apm{76.55}{0.49} \\
 & TBO & \apm{53.93}{0.83}$^*$ & \apm{36.00}{0.11} & \apm{69.50}{0.45}$^*$ & \apm{34.98}{0.16} \\
 & Ours + HX & \apm{\textbf{63.89}}{0.49}$^*$ & \apm{58.95}{0.09} & \apm{\textbf{78.81}}{0.42}$^*$ & \apm{\textbf{77.45}}{0.96} \\
 & Ours + TBO & \apm{44.36}{0.58} & \apm{35.95}{0.22} & \apm{63.19}{0.97}$^*$ & \apm{35.03}{0.90} \\
 & TBO + HX & \apm{47.74}{0.32} & \apm{56.14}{0.14} & \apm{55.67}{0.35} & \apm{64.61}{0.35}$^*$ \\
 & Ours + TBO + HX & \apm{40.95}{0.13} & \apm{55.47}{0.20}$^*$ & \apm{46.79}{0.66} & \apm{65.64}{0.23}$^*$ \\
\addlinespace
\multirow{7}{*}{RoBERTa} & Ours & \apm{57.68}{0.90}$^*$ & \apm{58.00}{14.7} & \apm{73.24}{0.81}$^*$ & \apm{71.25}{21.8} \\
 & HX & \apm{63.91}{0.76}$^*$ & \apm{60.32}{0.92} & \apm{\textbf{82.40}}{0.31} & \apm{\textbf{81.27}}{0.72} \\
 & TBO & \apm{55.56}{0.36}$^*$ & \apm{43.23}{1.34} & \apm{70.75}{0.31}$^*$ & \apm{48.54}{2.16} \\
 & Ours + HX & \apm{\textbf{64.58}}{0.99} & \apm{\textbf{63.91}}{0.82}$^\dagger$ & \apm{80.94}{0.62} & \apm{80.80}{1.18} \\
 & Ours + TBO & \apm{42.26}{0.81} & \apm{43.63}{0.23} & \apm{57.19}{2.13}$^*$ & \apm{48.73}{0.44} \\
 & TBO + HX & \apm{44.48}{2.36} & \apm{49.20}{1.07}$^*$ & \apm{50.21}{3.36} & \apm{59.25}{0.63}$^*$ \\
 & Ours + TBO + HX & \apm{36.93}{1.04} & \apm{52.47}{1.23}$^*$ & \apm{41.53}{0.32} & \apm{70.83}{0.82}$^*$ \\
\midrule
\multicolumn{2}{l}{Average per task} & 52.20 & 52.04 & 65.22 & 63.08 \\
\bottomrule
\end{tabular}

\caption{Median macro-F1$\pm$std (\%, 3 runs) for ALI vs ALC on IHC subsets (\textit{implicit}+neutral and \textit{explicit}+neutral).
\textit{Average per task} is the mean of the medians shown.
\textbf{Bold} text marks the best score for each approach within an abuse type and model; the $^*$ symbol is used to indicate that the corresponding score is significantly better in the ALI--ALC comparison for the same training setting, and the $^\dagger$ symbol is used to indicate that it is significantly better than the other training settings of the same model and approach.}
\label{median-macro-implicit-explicit-hate}
\end{center}
\end{table*}

\section{ALI vs ALC: full training configurations}
\label{annex:ali-vs-alc-full}
\label{annex:trust-ood-full}

Table~\ref{median-macro-ali-vs-classif-full} reports the exhaustive ALI vs ALC grid for all training settings, including combinations omitted from the focused main-text table (\textit{Ours}, \textit{HX}, \textit{TBO} only), on HatEval, IHC, JL-Hate, and the {TR}u{ST} human-annotated test split.
$^*$ marks the score that is significantly better in the ALI--ALC comparison for the same training setting, while $^\dagger$ marks the score that is significantly better than the other training settings of the same model and approach.

\begin{table*}[t]
    \begin{center}
    \resizebox{\textwidth}{!}{%
    \begin{tabular}{@{}l@{~}lc@{~}c@{~~}c@{~}c@{~~}c@{~}c@{~~}c@{~}c@{}}
    \toprule
    Models & Training sets & \multicolumn{2}{c}{HatEval} & \multicolumn{2}{c}{IHC} & \multicolumn{2}{c}{JL-Hate} & \multicolumn{2}{c}{{TR}u{ST}} \\
    \cmidrule(lr){3-4} \cmidrule(lr){5-6} \cmidrule(lr){7-8} \cmidrule(lr){9-10}
     & & ALI & ALC & ALI & ALC & ALI & ALC & ALI & ALC \\
    \midrule
    \multirow{7}{*}{fBERT} & Ours & \apm{54.64}{0.80}$^*$ & \apm{53.28}{0.11} & \apm{63.21}{0.83} & \apm{62.44}{0.26} & \apm{69.82}{0.81} & \apm{69.01}{0.67} & \apm{\textbf{71.97}}{0.43} & \apm{\textbf{74.22}}{0.07} \\
     & HX & \apm{\textbf{67.15}}{0.08}$^\dagger$ & \apm{\textbf{67.87}}{0.30}$^\dagger$ & \apm{68.00}{0.21}$^*$ & \apm{65.70}{0.86} & \apm{67.44}{0.57} & \apm{71.14}{3.98}$^*$ & \apm{68.34}{0.53} & \apm{70.62}{1.98} \\
     & TBO & \apm{50.06}{0.38}$^*$ & \apm{48.44}{1.64} & \apm{59.90}{3.21}$^*$ & \apm{42.55}{1.95} & \apm{69.88}{1.34}$^*$ & \apm{43.62}{0.55} & \apm{68.28}{0.21}$^*$ & \apm{43.35}{1.03} \\
     & Ours + HX & \apm{60.98}{0.54} & \apm{66.96}{0.14}$^*$ & \apm{\textbf{70.13}}{0.70}$^{*\dagger}$ & \apm{\textbf{65.82}}{0.31} & \apm{\textbf{74.84}}{0.49} & \apm{\textbf{76.37}}{0.54}$^\dagger$ & \apm{70.86}{0.03} & \apm{73.32}{0.39} \\
     & Ours + TBO & \apm{53.09}{0.31}$^*$ & \apm{48.14}{2.38} & \apm{44.62}{3.21}$^*$ & \apm{43.00}{1.29} & \apm{66.89}{0.49}$^*$ & \apm{43.39}{0.55} & \apm{61.54}{1.24} & \apm{42.23}{1.25} \\
     & TBO + HX & \apm{52.43}{0.47}$^*$ & \apm{50.91}{0.27} & \apm{38.59}{1.55} & \apm{54.23}{0.58}$^*$ & \apm{69.51}{1.23}$^*$ & \apm{47.03}{0.81} & \apm{50.85}{1.62} & \apm{61.87}{0.39}$^*$ \\
     & Ours + TBO + HX & \apm{51.22}{0.90} & \apm{53.73}{1.31}$^*$ & \apm{33.04}{1.74} & \apm{56.96}{1.61}$^*$ & \apm{55.09}{2.36} & \apm{56.13}{0.97} & \apm{44.64}{2.06} & \apm{64.48}{0.99}$^*$ \\
     \addlinespace
    \multirow{7}{*}{HateBERT} & Ours & \apm{55.84}{0.60}$^*$ & \apm{43.29}{5.26} & \apm{60.66}{0.75}$^*$ & \apm{48.83}{5.00} & \apm{70.26}{0.96}$^*$ & \apm{47.55}{9.28} & \apm{\textbf{73.83}}{0.53}$^*$ & \apm{58.90}{12.30} \\
     & HX & \apm{\textbf{61.80}}{1.07}$^\dagger$ & \apm{\textbf{67.17}}{0.26}$^{*\dagger}$ & \apm{68.72}{0.31}$^*$ & \apm{\textbf{65.59}}{0.16} & \apm{69.17}{1.97} & \apm{68.68}{0.84} & \apm{67.09}{0.63} & \apm{70.70}{0.65} \\
     & TBO & \apm{49.86}{1.18}$^*$ & \apm{39.05}{0.40} & \apm{59.46}{0.83}$^*$ & \apm{27.41}{0.11} & \apm{73.07}{0.14}$^*$ & \apm{41.12}{0.56} & \apm{71.21}{0.21}$^*$ & \apm{36.29}{0.28} \\
     & Ours + HX & \apm{56.66}{1.31} & \apm{65.09}{1.48}$^*$ & \apm{\textbf{69.26}}{0.49}$^*$ & \apm{64.91}{0.21} & \apm{\textbf{74.14}}{0.96} & \apm{\textbf{71.65}}{0.68} & \apm{71.60}{1.09} & \apm{\textbf{72.23}}{0.36} \\
     & Ours + TBO & \apm{53.06}{0.63}$^*$ & \apm{38.69}{0.07} & \apm{48.22}{0.90}$^*$ & \apm{27.40}{0.06} & \apm{66.19}{1.29}$^*$ & \apm{41.37}{0.29} & \apm{63.29}{0.38} & \apm{36.63}{0.06} \\
     & TBO + HX & \apm{51.52}{0.07}$^*$ & \apm{47.85}{0.08} & \apm{45.84}{0.28} & \apm{59.27}{0.27}$^*$ & \apm{71.65}{1.53}$^*$ & \apm{57.96}{0.21} & \apm{55.49}{1.29} & \apm{62.46}{0.62} \\
     & Ours + TBO + HX & \apm{51.56}{0.63}$^*$ & \apm{46.71}{0.11} & \apm{36.44}{0.36} & \apm{58.36}{0.25}$^*$ & \apm{53.71}{1.52} & \apm{57.64}{0.51} & \apm{45.53}{0.79} & \apm{63.27}{0.08}$^*$ \\
    \addlinespace
    \multirow{7}{*}{RoBERTa} & Ours & \apm{52.20}{0.68}$^*$ & \apm{47.20}{0.89} & \apm{64.38}{0.47} & \apm{63.16}{13.9} & \apm{71.05}{1.11} & \apm{72.29}{15.7} & \apm{\textbf{75.53}}{0.34} & \apm{71.57}{19.04} \\
     & HX & \apm{\textbf{65.58}}{0.86}$^\dagger$ & \apm{\textbf{65.06}}{1.48}$^\dagger$ & \apm{69.98}{0.58}$^*$ & \apm{66.82}{1.08} & \apm{71.51}{1.39}$^*$ & \apm{65.50}{2.20} & \apm{71.42}{0.94} & \apm{71.05}{1.16} \\
     & TBO & \apm{49.62}{1.23}$^*$ & \apm{43.55}{1.05} & \apm{61.13}{0.32}$^*$ & \apm{39.64}{2.22} & \apm{63.23}{1.88}$^*$ & \apm{36.56}{0.13} & \apm{72.92}{0.03}$^*$ & \apm{40.63}{1.53} \\
     & Ours + HX & \apm{58.03}{2.96} & \apm{61.81}{1.80}$^*$ & \apm{\textbf{70.48}}{1.07} & \apm{\textbf{70.21}}{1.02}$^\dagger$ & \apm{\textbf{78.27}}{1.03} & \apm{\textbf{75.21}}{0.43} & \apm{74.93}{0.21} & \apm{\textbf{74.66}}{0.86}$^\dagger$ \\
     & Ours + TBO & \apm{52.63}{0.28}$^*$ & \apm{43.48}{0.19} & \apm{42.82}{1.71}$^*$ & \apm{40.04}{0.34} & \apm{63.23}{1.88}$^*$ & \apm{36.56}{0.31} & \apm{58.23}{2.49} & \apm{41.23}{0.16} \\
     & TBO + HX & \apm{50.75}{0.61}$^*$ & \apm{49.39}{1.90} & \apm{40.28}{3.32} & \apm{47.64}{0.76}$^*$ & \apm{76.09}{1.39}$^*$ & \apm{51.73}{1.15} & \apm{57.35}{0.47} & \apm{62.26}{0.44} \\
     & Ours + TBO + HX & \apm{51.13}{0.21}$^*$ & \apm{47.28}{2.84} & \apm{31.13}{0.61} & \apm{57.32}{0.66}$^*$ & \apm{57.55}{3.17}$^*$ & \apm{47.15}{2.85} & \apm{45.98}{0.66} & \apm{61.89}{0.99} \\
    \midrule
    \multicolumn{2}{l}{Average per task} & 54.75 & 52.14 & 54.59 & 53.68 & 68.22 & 56.08 & 63.85 & 59.71 \\
    \bottomrule
    \end{tabular}%
    }
    \caption{Median macro-F1$\pm$std (\%, 3 runs) for ALI vs ALC on HatEval, IHC, JL-Hate, and {TR}u{ST} (all training combinations; the main text keeps \textit{Ours}/\textit{HX}/\textit{TBO} only; {TR}u{ST}: out of domain, $N=982$).
    \textit{Average per task} is the mean over the 21 configurations shown.
    \textbf{Bold} text marks the best score for each approach within a model and test set; the $^*$ symbol is used to indicate that the corresponding score is significantly better in the ALI--ALC comparison for the same training setting, and the $^\dagger$ symbol is used to indicate that it is significantly better than the other training settings of the same model and approach.}
    \label{median-macro-ali-vs-classif-full}
    \label{annex:trust-ood-grid}
    \end{center}
\end{table*}

\paragraph{Fine analysis of training combinations.}
Combining TBO with either corpus degrades ALI on IHC and HatEval, likely because TBO's targeted offensive spans sit outside the broader AL categories of those test sets (e.g., RoBERTa on IHC: $61.13$\% $\rightarrow$ $42.82$\%), whereas adding HX to our corpus improves ALI on IHC ($+6.1$ to $+8.6$; best: $70.48$\%) and JL-Hate (up to $78.27$\%) but can hurt HatEval (RoBERTa: $65.58$\% $\rightarrow$ $58.03$\%).
The same pattern holds on {TR}u{ST}: adding TBO to our corpus drops ALI sharply (fBERT $71.97$\% $\rightarrow$ $61.54$\%; RoBERTa $75.53$\% $\rightarrow$ $58.23$\%), the three-corpus mixture is the weakest ALI setting for every backbone ($44.64$--$45.98$\%), and combining our corpus with HX stays within one point of the best ALI ($70.86$--$74.93$\%).
Merging therefore helps when the added spans enrich the evaluated phenomenon and hurts when they introduce out-of-scope AL types; on the TBO mixtures, ALC reverses this ordering and gains from HX ($42.23$\% $\rightarrow$ $64.48$\% for fBERT), which suggests that sentence labels from HX rescue the sentence-level decision while token supervision is diluted.

\section{Span level obtained results}
\label{annex:span-level-obtained-results}

Table~\ref{annex:span-full-grid} reports the exhaustive span grid for all training configurations on JL-Hate (AL and target spans) and {TR}u{ST} (AL spans only),
with the same F1-exact / F1-partial / F1-token metrics and significance conventions as Table~\ref{table:span-level-obtained-results}.

\begin{table*}[t]
    \begin{center}
    \resizebox{\textwidth}{!}{
    \begin{tabular}{llccccccccc}
    \toprule
    \multirow{2}{*}{Models} & \multirow{2}{*}{Training sets} & \multicolumn{3}{c}{JL-Hate AL} & \multicolumn{3}{c}{JL-Hate Target} & \multicolumn{3}{c}{{TR}u{ST} AL} \\
    \cmidrule(lr){3-5} \cmidrule(lr){6-8} \cmidrule(lr){9-11}
     & & F1-exact & F1-partial & F1-token & F1-exact & F1-partial & F1-token & F1-exact & F1-partial & F1-token \\
    \midrule
\multirow{10}{*}{fBERT} & Ours & \apm{18.20}{0.40} & \apm{51.85}{0.34} & \apm{67.72}{0.55} & -- & -- & -- & \apm{13.62}{1.04} & \apm{61.23}{0.80} & \apm{48.53}{0.36} \\
 & Ours (with target) & \apm{17.49}{0.90} & \apm{51.75}{0.64} & \apm{67.66}{0.43} & \apm{\textbf{24.08}}{0.92}$^\dagger$ & \apm{\textbf{38.86}}{0.37}$^\dagger$ & \apm{\textbf{66.17}}{0.64}$^\dagger$ & \apm{13.70}{1.25} & \apm{59.45}{0.71} & \apm{48.99}{0.89} \\
 & HX & \apm{13.12}{0.68} & \apm{36.31}{0.12} & \apm{61.93}{0.16} & -- & -- & -- & \apm{17.68}{0.90} & \apm{52.46}{0.43} & \apm{46.66}{0.25} \\
 & TBO & \apm{\textbf{26.52}}{0.35} & \apm{51.61}{0.88} & \apm{\textbf{71.40}}{0.55} & -- & -- & -- & \apm{27.50}{0.88} & \apm{60.09}{0.57} & \apm{\textbf{49.31}}{0.78}$^\dagger$ \\
 & TBO (with target) & \apm{25.93}{0.48} & \apm{50.85}{0.44} & \apm{71.03}{0.48} & \apm{18.49}{1.13} & \apm{29.82}{0.78} & \apm{61.05}{0.15} & \apm{\textbf{27.58}}{1.66} & \apm{\textbf{62.15}}{0.70} & \apm{48.25}{0.30} \\
 & Ours + HX & \apm{19.37}{1.28} & \apm{\textbf{51.92}}{0.70} & \apm{68.32}{0.73} & -- & -- & -- & \apm{20.46}{0.20} & \apm{60.69}{0.89} & \apm{48.41}{0.34} \\
 & Ours + TBO & \apm{22.28}{1.15} & \apm{39.95}{0.93} & \apm{65.86}{0.27} & -- & -- & -- & \apm{22.20}{2.15} & \apm{41.73}{2.68} & \apm{41.15}{0.41} \\
 & Ours + TBO (with target) & \apm{23.95}{0.53} & \apm{45.45}{1.61} & \apm{67.87}{0.71} & \apm{21.78}{2.36} & \apm{34.42}{1.78} & \apm{63.08}{0.84} & \apm{24.43}{1.45} & \apm{42.75}{1.95} & \apm{41.20}{0.39} \\
 & TBO + HX & \apm{23.10}{0.40} & \apm{45.94}{1.29} & \apm{69.25}{1.06} & -- & -- & -- & \apm{13.16}{1.22} & \apm{25.38}{2.29} & \apm{38.84}{0.33} \\
 & Ours + TBO + HX & \apm{16.29}{2.62} & \apm{28.29}{2.82} & \apm{61.09}{1.12} & -- & -- & -- & \apm{7.98}{1.15} & \apm{15.41}{3.02} & \apm{37.65}{0.26} \\
\addlinespace
\multirow{10}{*}{HateBERT} & Ours & \apm{19.07}{0.83} & \apm{54.09}{1.08} & \apm{67.98}{0.82} & -- & -- & -- & \apm{12.81}{0.32} & \apm{62.74}{0.25} & \apm{48.01}{0.21} \\
 & Ours (with target) & \apm{19.35}{1.02} & \apm{\textbf{54.63}}{0.59} & \apm{68.41}{0.17} & \apm{\textbf{24.06}}{1.29} & \apm{\textbf{37.74}}{0.51} & \apm{\textbf{65.93}}{0.10} & \apm{16.33}{1.70} & \apm{\textbf{63.27}}{1.04} & \apm{48.33}{0.18} \\
 & HX & \apm{12.44}{1.05} & \apm{40.07}{1.10} & \apm{63.70}{0.58} & -- & -- & -- & \apm{19.14}{0.36} & \apm{55.13}{0.29} & \apm{48.29}{0.13} \\
 & TBO & \apm{\textbf{27.06}}{0.49} & \apm{52.61}{1.34} & \apm{\textbf{71.52}}{0.41} & -- & -- & -- & \apm{\textbf{28.18}}{1.49} & \apm{60.91}{0.37} & \apm{47.07}{0.64} \\
 & TBO (with target) & \apm{25.69}{1.62} & \apm{50.62}{0.93} & \apm{71.48}{0.42} & \apm{21.06}{2.16} & \apm{32.82}{1.62} & \apm{62.88}{0.89} & \apm{27.91}{0.52} & \apm{61.22}{0.38} & \apm{47.62}{0.24} \\
 & Ours + HX & \apm{15.92}{1.64} & \apm{51.35}{0.72} & \apm{68.22}{0.42} & -- & -- & -- & \apm{20.93}{0.63} & \apm{61.66}{1.17} & \apm{\textbf{49.58}}{0.08} \\
 & Ours + TBO & \apm{21.29}{1.47} & \apm{40.35}{2.02} & \apm{65.61}{1.07} & -- & -- & -- & \apm{24.44}{1.02} & \apm{44.37}{0.33} & \apm{41.48}{0.08} \\
 & Ours + TBO (with target) & \apm{19.40}{0.73} & \apm{37.26}{2.01} & \apm{63.89}{1.09} & \apm{21.20}{0.32} & \apm{35.38}{1.02} & \apm{63.83}{0.42} & \apm{21.37}{2.73} & \apm{39.46}{3.28} & \apm{41.07}{0.50} \\
 & TBO + HX & \apm{25.15}{0.13} & \apm{48.19}{1.26} & \apm{69.84}{0.42} & -- & -- & -- & \apm{14.97}{0.79} & \apm{32.88}{1.97} & \apm{40.61}{0.36} \\
 & Ours + TBO + HX & \apm{15.92}{0.81} & \apm{27.10}{1.48} & \apm{59.92}{0.35} & -- & -- & -- & \apm{8.73}{0.76} & \apm{16.10}{1.62} & \apm{37.78}{0.16} \\
\addlinespace
\multirow{10}{*}{RoBERTa} & Ours & \apm{19.22}{2.79} & \apm{53.95}{1.28} & \apm{69.04}{1.37} & -- & -- & -- & \apm{15.19}{0.77} & \apm{64.75}{0.61} & \apm{50.41}{0.66} \\
 & Ours (with target) & \apm{22.53}{0.68} & \apm{56.67}{1.05} & \apm{68.16}{0.48} & \apm{31.13}{0.21} & \apm{43.35}{0.78} & \apm{\textbf{70.45}}{0.58} & \apm{15.89}{0.40} & \apm{\textbf{66.00}}{0.35} & \apm{\textbf{51.17}}{0.49} \\
 & HX & \apm{14.88}{0.51} & \apm{42.29}{0.78} & \apm{65.06}{0.45} & -- & -- & -- & \apm{18.76}{1.66} & \apm{57.20}{0.44} & \apm{49.05}{0.11} \\
 & TBO & \apm{27.67}{0.60} & \apm{54.10}{0.34} & \apm{71.71}{0.31} & -- & -- & -- & \apm{27.96}{1.52} & \apm{60.82}{1.24} & \apm{48.79}{0.22} \\
 & TBO (with target) & \apm{27.89}{0.87} & \apm{53.52}{0.36} & \apm{\textbf{72.55}}{0.18} & \apm{30.61}{0.81} & \apm{43.29}{1.11} & \apm{68.03}{0.43} & \apm{\textbf{28.34}}{1.00} & \apm{60.59}{0.92} & \apm{48.29}{0.57} \\
 & Ours + HX & \apm{22.44}{2.09} & \apm{\textbf{58.25}}{1.67} & \apm{71.39}{1.58} & -- & -- & -- & \apm{21.25}{0.93} & \apm{65.02}{1.60} & \apm{50.42}{0.59} \\
 & Ours + TBO & \apm{18.90}{0.27} & \apm{35.20}{2.76} & \apm{64.52}{0.66} & -- & -- & -- & \apm{19.63}{1.17} & \apm{34.58}{4.39} & \apm{40.78}{0.62} \\
 & Ours + TBO (with target) & \apm{20.05}{2.40} & \apm{35.36}{3.54} & \apm{64.15}{1.59} & \apm{\textbf{32.30}}{0.78} & \apm{\textbf{43.48}}{0.62} & \apm{69.40}{0.65} & \apm{18.95}{0.63} & \apm{37.19}{2.16} & \apm{40.62}{0.35} \\
 & TBO + HX & \apm{\textbf{29.19}}{1.17}$^\dagger$ & \apm{53.98}{1.47} & \apm{72.30}{0.86} & -- & -- & -- & \apm{17.77}{0.20} & \apm{36.82}{1.48} & \apm{41.67}{0.51} \\
 & Ours + TBO + HX & \apm{16.95}{3.01} & \apm{29.60}{4.32} & \apm{62.06}{1.81} & -- & -- & -- & \apm{7.70}{1.29} & \apm{17.66}{1.61} & \apm{38.03}{0.21} \\
    \bottomrule
    \end{tabular}
    }
    \caption{Median F1$\pm$std (\%, 3 runs) for AL and target spans on JL-Hate and AL spans on {TR}u{ST} (all training settings).
    \textbf{Bold} text marks the best score for each F1 type within a backbone, and the $^\dagger$ symbol is used to indicate that the corresponding score is significantly better than the other scores of the same backbone.}
    \label{annex:span-full-grid}
    \end{center}
\end{table*}

\paragraph{Fine analysis of paired significance.}
Significant standard vs \textit{with target} differences are configuration-specific:
fBERT favors standard \textit{Ours} on all three AL metrics; HateBERT favors \textit{Ours (with target)} but standard \textit{TBO}; RoBERTa favors \textit{Ours (with target)} on F1-exact and F1-partial while standard \textit{Ours} remains higher on F1-token.
No significant paired difference is observed for \textit{Ours + TBO} across backbones.
For Target exact, \textit{Ours (with target)} is significantly better than \textit{TBO (with target)} for fBERT and HateBERT ($^\dagger$ for fBERT under the best-vs-all criterion), while the RoBERTa gap is not significant.
These patterns support a backbone-dependent effect of target supervision on AL span detection, while target-span gains are more consistent for Ours-vs-TBO in two out of three backbones.

\end{document}